\documentclass[10pt,journal]{IEEEtran}
\usepackage{amsmath,amsfonts}
\usepackage{algorithm}
\usepackage{algorithmicx}
\usepackage{algpseudocode}
\usepackage{array}
\usepackage[caption=false,font=footnotesize,labelfont=sf,textfont=sf]{subfig}
\usepackage{textcomp}
\usepackage{stfloats}
\usepackage{placeins}
\usepackage{url}
\usepackage{graphicx}
\graphicspath{{Figs/}}
\usepackage{cite}
\usepackage{color}
\usepackage{hyperref}
\hypersetup{pdfborder={0 0 0}}

\begin{document}

\title{Task-Oriented Visual Feature Compression via Residual Vector Quantization for Device-Edge Multimodal Inference}

\author{
Luning~Pang,
Cheng~Yuan,
Jiawei~Shao,~\IEEEmembership{Member,~IEEE},
Mingtao~Huang,
and~Yuan~Shen,~\IEEEmembership{Senior Member,~IEEE}%
\thanks{Luning Pang, Mingtao Huang, and Yuan Shen are with the
Department of Electronic Engineering, and Beijing National
Research Center for Information Science and Technology,
Tsinghua University, Beijing 100084, China
(e-mail: pln25@mails.tsinghua.edu.cn;
huangmt@mail.tsinghua.edu.cn;
shenyuan\_ee@tsinghua.edu.cn).
Corresponding author: Yuan Shen.}%
\thanks{Luning Pang, Cheng Yuan, and Jiawei Shao are with the
Institute of Artificial Intelligence (TeleAI), China Telecom,
Beijing 100033, China
(e-mail: pln25@mails.tsinghua.edu.cn;
yuanc3@chinatelecom.cn;
shaojw2@chinatelecom.cn).}%
}

\maketitle
\begin{abstract}
Large multimodal models (LMMs) support diverse visual understanding and reasoning tasks but are often impractical to run entirely on resource-constrained devices. Device--edge co-inference reduces device computation, yet transmitting visual data over bandwidth-limited uplinks can introduce substantial delay. Task-oriented feature compression (TOFC) reduces the payload through feature aggregation and entropy coding. However, continuous-feature coding remains costly, and query-agnostic aggregation may discard task-relevant local evidence. We propose query-guided task-oriented feature compression (Q-TOFC) for device--edge multimodal inference. Q-TOFC employs residual vector quantization (RVQ) to encode each merged feature as a compact sequence of codebook indices, reducing its representation cost and allowing more features to be transmitted. It further incorporates query relevance into feature aggregation and uses a quantization error compensation adapter to mitigate the distortion introduced by discrete quantization. Experiments on seven multimodal benchmarks show that Q-TOFC reduces the visual payload by 53.6\% relative to TOFC while maintaining comparable average normalized task performance. End-to-end latency evaluations further demonstrate lower latency under bandwidth-constrained uplinks.
\end{abstract}

\begin{IEEEkeywords}
Device-edge co-inference, feature compression, large multimodal models, residual vector quantization, task-oriented communication.
\end{IEEEkeywords}

\section{Introduction}

\begin{figure*}[!t]
    \centering
    \includegraphics[width=\linewidth]{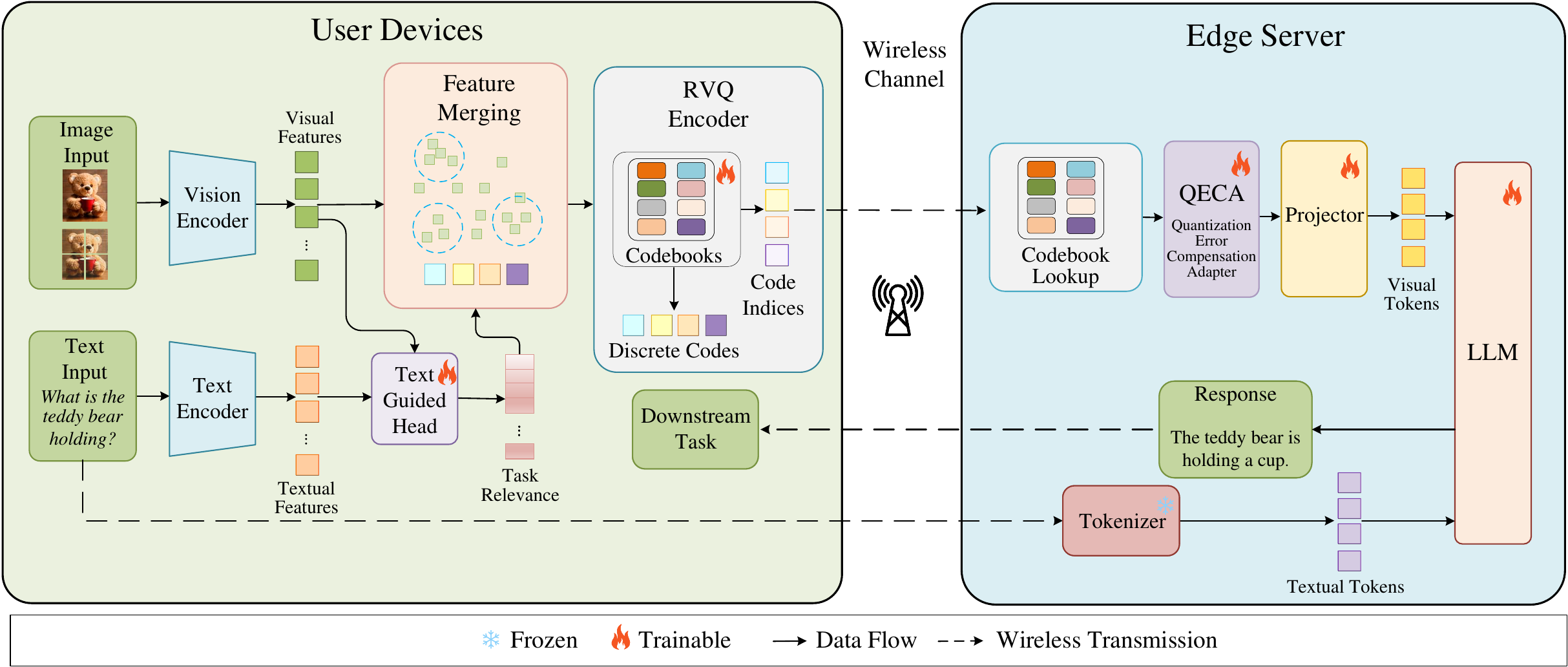}
    \caption{Overview of Q-TOFC. On the user device, the vision and text encoders extract visual and query features. A lightweight query-aware scoring module estimates feature relevance, after which query-guided clustering produces a compact set of merged features. Multi-layer residual vector quantization (RVQ) maps each merged feature to discrete indices. On the edge server, codebook lookup and a quantization error compensation adapter (QECA) reconstruct the features for subsequent LMM inference.
    }
    \label{fig_pipeline}
\end{figure*}

Large multimodal models (LMMs) combine visual perception with language reasoning \cite{LMMsurvey1} and support applications ranging from robotic perception \cite{LMMembodied} and autonomous navigation \cite{LMMdriving} to conversational agents \cite{LMMagent}. A typical LMM contains a vision encoder, such as CLIP \cite{CLIP} or SigLIP \cite{SigLIP}, a multimodal projector, and a large language model (LLM) that generates responses from visual and textual tokens \cite{llava,llava-ov,qwen2.5vl}. Running this entire pipeline on a resource-constrained mobile device is often impractical \cite{TOCsurvey2,TOCmagazineshao}. Device--edge co-inference therefore partitions the computation between the device and a nearby edge server \cite{aiflow,aiflow_perspectives}, but the partition also determines what visual data must traverse the wireless uplink.

Visual information can be offloaded at either the image or feature level. Image-level offloading sends a compressed image to the edge server, where the image is decoded and processed by the vision encoder. Conventional JPEG \cite{wallace1991jpeg} and learned neural codecs \cite{Balle2017,Balle2018,Cheng2020,ELIC} can substantially reduce the image bitstream, but they leave image decoding and visual feature extraction at the server. Their compression objectives operate in the image domain and are designed to preserve reconstructed image quality. Consequently, they do not directly exploit redundancy in the intermediate visual representations consumed by the LMM.

Feature-level co-inference instead executes the vision encoder on the device and transmits its intermediate representations, avoiding visual feature extraction at the server. The obstacle is the size of dense visual features.
For the LLaVA-OneVision backbone used in our primary experiments, one SigLIP patch contains 729 features of dimension 1152 and exceeds 1.6\,MiB in FP16.
High-resolution inputs may contain multiple patches. Task-oriented feature compression (TOFC) \cite{TOFC} demonstrates that directly compressing these representations can provide a stronger communication--performance trade-off than image-level codecs.
TOFC first applies density peaks clustering based on $K$ nearest neighbors (DPC-KNN) to aggregate the visual features in each patch into a smaller set of merged representations, thereby reducing the number of features to be transmitted.
It then entropy-codes the merged representations.

TOFC establishes the value of feature-level transmission, but its compression mechanism leaves two opportunities for improvement under a tight uplink budget. First, DPC-KNN selects cluster centers from the visual feature distribution without using the textual query, even though the evidence needed by the downstream LMM is query dependent. Second, very low payloads require reducing each patch to a small number of merged features. As fewer representations must summarize the original visual tokens, localized evidence such as a sign or small object can be lost. These observations motivate a different allocation of the communication budget: retain more query-relevant merged features while reducing the representation cost of each retained feature.

We propose query-guided task-oriented feature compression (Q-TOFC), a visual feature compression framework that combines query-aware token reduction with RVQ-based coding and feature refinement. A lightweight relevance score incorporates the textual query into the DPC-KNN center-ranking criterion. The total number of selected centers remains unchanged, but visual features that are more relevant to the query are more likely to be selected as cluster centers. Q-TOFC adapts residual vector quantization (RVQ) \cite{soundstream,encodec} to encode each merged visual feature as a compact sequence of codebook indices. Reducing the representation cost per feature allows the framework to retain more merged features at a low overall payload. The edge server reconstructs the features through codebook lookup, after which a quantization error compensation adapter (QECA) predicts a residual correction for the quantized representation. During end-to-end fine-tuning, QECA is jointly shaped by the language-modeling objective and feature-level VQ regularization, balancing downstream task supervision with feature fidelity.

We compare Q-TOFC with TOFC, ELIC, and JPEG on seven multimodal benchmarks using rate--performance and end-to-end latency evaluations. At comparable average normalized performance, Q-TOFC reduces the visual payload by 53.6\% relative to TOFC. Under bandwidth-constrained uplinks, the lower payload also translates into lower end-to-end latency despite the additional device-side encoding operations. Additional experiments with LLaVA-1.5 show that the same trend persists with a different vision encoder and image-patching scheme.

Our main contributions are summarized as follows:
\begin{itemize}
    \item We propose a query-conditioned visual feature compression framework for device--edge LMM inference, where DPC-KNN density--diversity scores are combined with text--visual relevance to guide cluster-center selection.
    \item We design an RVQ-based codec that lowers the representation cost of each merged feature, enabling more query-relevant features to be retained under a constrained payload, and introduce QECA to refine the reconstructed representations.
    \item We validate Q-TOFC through extensive experiments against TOFC, ELIC, and JPEG on seven multimodal benchmarks. Q-TOFC reduces the visual payload by 53.6\% relative to TOFC while maintaining comparable average normalized performance. In the primary LLaVA-OneVision setting, it further reduces end-to-end latency by up to 29.0\% relative to TOFC under bandwidth-constrained uplinks.
    \item Component ablations show that query guidance and QECA provide complementary performance gains. The RVQ-depth study further characterizes the trade-off among communication cost, codebook storage, and task performance.
\end{itemize}

The remainder of the paper is organized as follows.
Section~\ref{sec:related} surveys related literature.
Section~\ref{sec:system} formalizes the system model.
Section~\ref{sec:method} details the Q-TOFC framework.
Section~\ref{sec:experiment} reports experimental results, and Section~\ref{sec:conclusion} concludes the paper.

\section{Related Work}
\label{sec:related}

\subsection{Task-Oriented Communication}
Conventional wireless communication systems treat every bit of the transmitted payload equally, aiming for faithful reconstruction of the source signal.
Task-oriented communication departs from this philosophy by asking a different question: what information does the receiver actually need to perform well on its assigned task?
The theoretical foundation is often traced to the information bottleneck (IB) principle \cite{tishby2000information}, which formalizes the trade-off between compression and relevance.
Practical instantiations of this idea have been explored in feature compression for device-edge inference \cite{shao2020bottlenet++}, cooperative edge inference \cite{shao2022task}, robust task-oriented transmission \cite{xie2023robust}, and video analytics over temporally correlated frames \cite{TOCvideo}.
Task-oriented edge--cloud co-inference has also been extended to human action understanding by converting pose sequences into discrete motion tokens for transmission \cite{liu2026action}.
A common theme is that discarding task-irrelevant redundancy at the transmitter can shrink the payload by more than an order of magnitude with little impact on accuracy.
TOFC \cite{TOFC} applies this principle to device--edge LMM inference.
It places vision encoding and feature compression on the user device, while the edge server decodes the transmitted representation, applies the multimodal projector, and performs LLM inference.

\subsection{Neural Data Compression}
Reducing the storage and transmission cost of images has a long history.
Hand-crafted codecs such as JPEG \cite{wallace1991jpeg} and JPEG 2000 \cite{skodras2001jpeg} apply block-based frequency transforms followed by quantization and entropy coding.
Over the past decade, learned compression methods have pushed rate--distortion curves well beyond these classical bounds.
Ball\'{e} et al.\ \cite{Balle2017, Balle2018} introduced the hyperprior architecture that jointly optimizes a nonlinear transform with a learned probability model, and subsequent work added autoregressive context modeling \cite{Cheng2020, minnen2018joint, checkerboard} and more expressive prior structures \cite{ELIC}.
Despite their success, these methods are generally trained to minimize pixel-level or perceptual distortion metrics such as PSNR, MS-SSIM, and LPIPS. Their bit allocation may therefore be inefficient when the downstream consumer is an LMM rather than a human viewer.
In contrast, the present work compresses the \emph{latent features} produced by the vision encoder and jointly optimizes the compression modules using the downstream token-prediction loss and feature-level VQ losses.

\subsection{Vector Quantization for Representation Learning}
Vector quantization (VQ) replaces a continuous vector with the nearest entry in a learned codebook, yielding a discrete, compact code.
The seminal VQ-VAE \cite{vqvae} demonstrated that such discretized representations can still support high-fidelity generation when coupled with an appropriate decoder.
In the audio domain, SoundStream \cite{soundstream} and EnCodec \cite{encodec} introduced RVQ, which stacks multiple VQ layers: after the first layer quantizes the input, each subsequent layer quantizes the residual between the input and the sum of previously selected codewords.
This coarse-to-fine strategy achieves high spectral fidelity with a small number of bits per frame.
In the visual domain, VQGAN \cite{VQGAN} combined VQ with a perceptual loss and a patch-wise discriminator to learn codebooks for image synthesis.
We repurpose RVQ as a compression primitive for task-oriented visual feature transmission and couple it with a lightweight compensation adapter for downstream LMM inference.

\subsection{LMM Inference Acceleration}
Processing hundreds of visual tokens through dozens of transformer layers is a major source of latency in LMM inference, motivating extensive research on visual token reduction.
Attention-based methods such as FastV and MustDrop remove low-importance visual tokens during LLM inference \cite{FastV,MustDrop}, while projector-side methods such as TokenPacker and LLaVA-Mini condense the vision-encoder output before it enters the LLM \cite{TokenPacker,LLaVA-Mini}.
Other methods exploit redundancy among visual tokens.
LLaVA-PruMerge adaptively prunes and merges tokens, VisionZip retains dominant tokens together with contextual summaries, and FocusLLaVA adjusts visual granularity through coarse-to-fine selection \cite{LLaVA-PruMerge,VisionZip,FocusLLaVA}.

TOFC \cite{TOFC} places token reduction on the user device before feature transmission.
It uses DPC-KNN to group visual features and averages the features assigned to each cluster.
Consequently, features that are not selected as cluster centers still contribute to the merged representations instead of being directly discarded.
The resulting smaller feature set reduces both the communication cost and the number of visual tokens processed by the server-side LMM.

Q-TOFC builds on this merge-before-transmission design.
The main difference lies in how the merged features are represented.
TOFC applies learned entropy coding after reducing each patch to a small set of merged features, whereas Q-TOFC uses RVQ to encode each merged feature as a compact sequence of codebook indices.
The lower representation cost per feature allows Q-TOFC to retain more merged features under a constrained communication budget.

\section{System Model}
\label{sec:system}

We consider a device-edge co-inference system in which a resource-constrained user device communicates with an edge server over a wireless uplink channel, as depicted in Fig.~\ref{fig_pipeline}.
The user submits a multimodal query consisting of an image $\boldsymbol{v}_{\rm i} \in \mathbb{R}^{h \times w \times 3}$ and a textual instruction in natural language.
Because the instruction is short (tens of tokens), it is available to the on-device compression module and is also sent directly to the server without compression.
The image, which dominates the transmission budget, is processed locally by the device as described below.

Following the design of recent LMMs \cite{llava-ov, llava}, the backbone preprocessing scheme converts the input image into $n_{\rm p}$ encoder inputs, such as a single resized image or multiple image patches.
The vision encoder $f_{\rm vis}$ transforms these inputs, denoted by $\boldsymbol{v}_{\rm p}$, into a tensor of visual features:
\begin{equation}
    \boldsymbol{X} = f_{\rm vis}(\boldsymbol{v}_{\rm p}),\quad \boldsymbol{X} \in \mathbb{R}^{n_{\rm p} \times n_{\rm v} \times d_{\rm v}},
    \label{eq:vision_enc}
\end{equation}
where $n_{\rm v}$ and $d_{\rm v}$ denote the number of features per patch and the feature dimension, respectively.

The raw visual features $\boldsymbol{X}$ are far too large for a bandwidth-limited uplink.
We therefore apply a cascade of lightweight modules on the device:
(i) a query-aware token scoring module that estimates the relevance of each visual feature to the textual query;
(ii) a query-guided clustering module that groups the $n_{\rm v}$ features in each patch into $n_{\rm c}$ clusters and averages the features within each cluster to produce $n_{\rm c}$ merged features; and
(iii) an RVQ codec that maps each merged feature to a sequence of codebook indices.
The resulting indices are transmitted to the edge server.

Upon receiving the indices, the edge server reconstructs the quantized features through codebook lookup and applies QECA, detailed in Section~\ref{sec:method}, to reduce the distortion introduced by quantization.
The reconstructed features are projected into the word-embedding space of the LLM by a two-layer MLP and concatenated with the tokenized instruction.
The LLM then autoregressively generates the textual response, which is sent back to the user device.

Following the convention in task-oriented communication research \cite{TOCvideo, shao2022task}, our channel model focuses on the uplink bandwidth constraint rather than explicitly modeling physical-layer effects such as fading and modulation, as channel coding and signal processing are beyond the scope of this work.

\subsection{Communication Cost Model}
\label{sec:cost_model}

We characterize the communication cost by the number of bits used to transmit the visual representation for each multimodal query. Because all methods transmit the same textual instruction, it is not included in the visual payload. If the device transmits the visual features produced by the vision encoder without compression, the
payload is
\begin{equation}
    B_{\rm raw}=n_{\rm p}n_{\rm v}d_{\rm v}b_{\rm f},
    \label{eq:raw_bits}
\end{equation}
where $b_{\rm f}$ is the number of bits used to represent each feature element.

For an $L$-layer RVQ, each layer produces one codebook index for every merged feature. Let $M_k$ denote the number of entries in the codebook at layer $k$. The index produced by that layer requires $\lceil\log_2 M_k\rceil$ bits. The number of index bits required for one merged feature is therefore
\begin{equation}
    B_{\rm tok}
    =
    \sum_{k=0}^{L-1}\left\lceil\log_2 M_k\right\rceil.
    \label{eq:rvq_bits_per_feature}
\end{equation}
The RVQ index payload of one request is
\begin{equation}
    B_{\rm Q}=n_{\rm p}n_{\rm c}B_{\rm tok},
    \label{eq:qtofc_bits}
\end{equation}
where $n_{\rm c}$ is the number of merged features per patch. The payload relative to uncompressed feature transmission is
\begin{equation}
    \frac{B_{\rm Q}}{B_{\rm raw}}
    =
    \frac{n_{\rm c}B_{\rm tok}}
    {n_{\rm v}d_{\rm v}b_{\rm f}}.
    \label{eq:payload_ratio}
\end{equation}
The communication cost is therefore determined by the patch count $n_{\rm p}$, the number of merged features $n_{\rm c}$, and the RVQ configuration through $B_{\rm tok}$.

Let $R_{\rm u}$ denote the effective uplink throughput in bits per
second. The transmission latency is
\begin{equation}
    T_{\rm tx}=\frac{B_{\rm Q}}{R_{\rm u}}.
    \label{eq:tx_latency}
\end{equation}
For a selected value of $n_{\rm c}$ and a given RVQ configuration, the transmission latency follows from the image patch count and the uplink throughput.
The end-to-end latency considered in our evaluation is
\begin{equation}
    T_{\rm e2e}=T_{\rm dev}+T_{\rm tx}+T_{\rm edge},
    \label{eq:e2e_latency}
\end{equation}
where $T_{\rm dev}$ includes on-device visual processing and compression, and $T_{\rm edge}$ includes reconstruction and the remaining LMM computation. This decomposition is important because a smaller payload can require additional device computation, while image-level codecs move vision encoding to the server. We therefore report all three components rather than using payload size as a proxy for end-to-end latency.

\subsection{Design Objective}
\label{sec:objective}

The goal of Q-TOFC is not solely to minimize reconstruction error with respect to the original image or visual features.
Instead, the codec should preserve the information required by the downstream LMM to answer the user's query.
Let $\mathcal{D}$ denote the distribution of image--query--answer tuples, and let $\mathcal{C}_\phi(\cdot,\cdot)$ denote the query-conditioned compression and reconstruction pipeline parameterized by $\phi$.
The ideal task-oriented objective can be written as
\begin{equation}
\begin{aligned}
    \min_{\phi}\quad &
    \mathbb{E}_{\mathcal{D}}\!\left[
        \mathcal{L}_{\rm task}\!\left(
        F_\theta(\mathcal{C}_\phi(f_{\rm vis}(\boldsymbol{v}_{\rm i}),\boldsymbol{q}),\boldsymbol{q}),
        \boldsymbol{a}\right)\right] \\
    \text{s.t.}\quad & B_{\rm Q} \le B_0,
\end{aligned}
    \label{eq:constrained_objective}
\end{equation}
where $F_\theta$ is the server-side LMM, $\boldsymbol{q}$ is the textual instruction, $\boldsymbol{a}$ is the target answer, and $B_0$ is the communication budget.
This formulation makes two requirements explicit.
First, token reduction should be conditioned on the query because relevance is task dependent.
Second, the reconstruction module should be optimized through the LMM loss rather than solely through feature-level distortion.
The following section instantiates these two requirements with query-guided clustering and QECA.

\section{Method}
\label{sec:method}

This section presents the proposed Q-TOFC framework and its three key components.
We first describe the query-guided clustering module that reduces the number of visual features while biasing cluster-center selection toward query-relevant evidence (Section~\ref{sec:cluster}), then introduce the RVQ codec (Section~\ref{sec:rvq}), followed by QECA (Section~\ref{sec:qeca}), and finally discuss the training strategy (Section~\ref{sec:training}).

\subsection{Query-Guided Token Reduction}
\label{sec:cluster}

Modern vision encoders produce hundreds of features per image patch, many of which lie close together in the feature space and carry redundant information.
Following the feature-merging design of TOFC \cite{TOFC}, we adopt density peaks clustering based on $K$ nearest neighbors (DPC-KNN) \cite{DPC-KNN}, a training-free method that adaptively partitions the $n_{\rm v}$ features in each patch into $n_{\rm c}$ clusters based on their distribution in the latent space.
The DPC-KNN density--diversity criterion is inherited from this prior design. Our modification is to incorporate query relevance into the center-ranking score.
Because the textual query is available before compression, we further introduce a lightweight query-aware scoring mechanism that biases the cluster-center selection toward task-relevant visual tokens.

For the $n$-th patch, the local density of the $i$-th feature is computed from its $K$ nearest neighbors as
\begin{equation}
    \rho_{n,i} = \exp\!\Bigl(-\frac{1}{K} \sum_{\boldsymbol{x}_{n,j} \in {\rm KNN}(\boldsymbol{x}_{n,i},\,K)} \|\boldsymbol{x}_{n,i} - \boldsymbol{x}_{n,j}\|^2\Bigr).
    \label{eq:rho}
\end{equation}
To encourage diversity among cluster centers in the feature space, the minimum distance between feature $i$ and any feature with higher density is also computed:
\begin{equation}
    \delta_{n,i} = \min_{j:\,\rho_{n,j}>\rho_{n,i}} \|\boldsymbol{x}_{n,i} - \boldsymbol{x}_{n,j}\|.
    \label{eq:delta}
\end{equation}
The $n_{\rm c}$ features with the largest $\rho_{n,i} \times \delta_{n,i}$ are selected as cluster centers, and the remaining features are assigned to their nearest center.
Average pooling within each cluster yields the merged features $\boldsymbol{Y} \in \mathbb{R}^{n_{\rm p} \times n_{\rm c} \times d_{\rm v}}$.

To condition center selection on the query, we use the text tower paired with the vision encoder to embed the user instruction. Contrastively pretrained vision--language encoders such as CLIP and SigLIP align their image and text representations in a shared multimodal space \cite{CLIP,SigLIP}. These representations therefore provide a natural basis for estimating text--visual relevance. Given the embedded text tokens $\boldsymbol{T} \in \mathbb{R}^{N_{\rm txt} \times d_{\rm txt}}$, the scoring module computes a relevance score for each visual feature:
\begin{equation}
    r_{n,i} = \max_{j}\, \operatorname{sim}\bigl(f_{\rm img}(\boldsymbol{x}_{n,i}),\; f_{\rm txt}(\boldsymbol{t}_j)\bigr),
    \label{eq:text_relevance}
\end{equation}
where $f_{\rm img}$ and $f_{\rm txt}$ are lightweight projection heads that map the image and text features into a shared low-dimensional relevance space, and $\operatorname{sim}(\cdot,\cdot)$ denotes cosine similarity.
The maximum over text tokens lets each visual token match the most relevant textual token in the query, which is useful for localized tasks such as OCR, counting, and object-centric reasoning.
We normalize the relevance scores within each patch as
\begin{equation}
    \tilde{r}_{n,i}=\frac{r_{n,i}-\min_j r_{n,j}}{\max_j r_{n,j}-\min_j r_{n,j}+\epsilon}.
    \label{eq:normalized_relevance}
\end{equation}
The clustering score is then adjusted as
\begin{equation}
    \hat{s}_{n,i} = s_{n,i} \cdot (1 + \alpha \tilde{r}_{n,i}),
    \label{eq:adjusted_score}
\end{equation}
where $s_{n,i} = \rho_{n,i} \times \delta_{n,i}$ is the original DPC-KNN score and $\alpha$ is a hyperparameter controlling the strength of text guidance.
When $\alpha = 0$, the clustering reduces to the standard DPC-KNN formulation.
The query-aware score preserves the density--diversity criterion of DPC-KNN through $s_{n,i}$ while biasing the selected cluster centers toward features with higher text relevance.

\subsection{Residual Vector Quantization Codec}
\label{sec:rvq}

After clustering, each merged feature vector $\boldsymbol{y} \in \mathbb{R}^{d_{\rm v}}$ must be represented compactly.
Q-TOFC uses the established RVQ formulation \cite{soundstream, encodec} to represent each merged feature as a sequence of discrete codebook indices through a coarse-to-fine approximation.
The discrete codec is combined with query-guided clustering and task-driven quantization compensation for task-oriented visual feature transmission.

Let $\{\boldsymbol{C}^{(k)}\}_{k=0}^{L-1}$ be a set of $L$ learnable codebooks, where $\boldsymbol{C}^{(k)} \in \mathbb{R}^{M_k \times d_{\rm v}}$ contains $M_k$ entries at layer $k$.
Given an input vector $\boldsymbol{y}$, our cosine-similarity implementation proceeds iteratively:
\begin{align}
    \boldsymbol{r}^{(0)} &= \boldsymbol{y}, \nonumber \\
    \boldsymbol{q}^{(k)} &= \arg\max_{\boldsymbol{c} \in \boldsymbol{C}^{(k)}}
    \operatorname{cos}(\boldsymbol{r}^{(k)},\boldsymbol{c}), \quad k = 0, 1, \ldots, L-1, \label{eq:rvq_nearest} \\
    \boldsymbol{r}^{(k+1)} &= \boldsymbol{r}^{(k)} - \boldsymbol{q}^{(k)},
    \label{eq:rvq_residual}
\end{align}
where $\boldsymbol{r}^{(k)}$ is the residual at layer $k$, and $\boldsymbol{q}^{(k)}$ is the selected codeword.
The reconstructed vector is the sum of all codewords:
\begin{equation}
    \hat{\boldsymbol{y}}_{\rm rvq} = \sum_{k=0}^{L-1} \boldsymbol{q}^{(k)}.
    \label{eq:rvq_sum}
\end{equation}
The index at layer $k$ requires $\lceil \log_2 M_k \rceil$ bits. Summing the index lengths across all layers gives $B_{\rm tok}$ in \eqref{eq:rvq_bits_per_feature}.

For efficient codebook lookup, \eqref{eq:rvq_nearest} is implemented as batched matrix multiplication after vector normalization.
This choice emphasizes angular alignment between normalized feature and codeword representations and is consistent with the cosine-similarity formulation in \eqref{eq:rvq_nearest}.

During training, the non-differentiable codeword selection operation is bypassed with the straight-through estimator (STE) \cite{bengio2013ste}: gradients are copied from the output directly to the input, enabling end-to-end backpropagation through the quantization bottleneck.
The codebook at each RVQ layer is updated using exponential moving averages of the assignment counts and the residual vectors assigned to each entry \cite{vqvae,soundstream}.

\begin{algorithm}[!t]
\caption{On-device Q-TOFC encoding}
\label{alg:qtofc_encoding}
\begin{algorithmic}[1]
\Require Image $\boldsymbol{v}_{\rm i}$, text query $\boldsymbol{q}$, number of merged features per patch $n_{\rm c}$, RVQ codebooks $\{\boldsymbol{C}^{(k)}\}_{k=0}^{L-1}$
\Ensure Ordered RVQ index sequence
\State Extract visual features $\boldsymbol{X}=f_{\rm vis}(\boldsymbol{v}_{\rm i})$
\State Extract text features $\boldsymbol{T}$ from the frozen text encoder
\For{$n=1,\ldots,n_{\rm p}$}
    \State Compute DPC-KNN density scores $\rho_{n,i}$ and diversity scores $\delta_{n,i}$
    \State Compute query relevance scores $r_{n,i}$ using \eqref{eq:text_relevance}
    \State Select $n_{\rm c}$ cluster centers by the adjusted score in \eqref{eq:adjusted_score}
    \State Average-pool features within each cluster to obtain $\{\boldsymbol{y}_{n,m}\}_{m=1}^{n_{\rm c}}$
    \For{$m=1,\ldots,n_{\rm c}$}
        \State Quantize $\boldsymbol{y}_{n,m}$ through $L$ RVQ layers using \eqref{eq:rvq_nearest}--\eqref{eq:rvq_residual}
        \State Append the $L$ fixed-width RVQ indices to the output sequence
    \EndFor
\EndFor
\State \Return ordered RVQ index sequence
\end{algorithmic}
\end{algorithm}

\subsection{Quantization Error Compensation Adapter}
\label{sec:qeca}

Multi-layer RVQ inevitably introduces quantization error: the sum of $L$ codewords only approximates the original continuous vector.
Because this approximation error can degrade downstream inference, we propose QECA to recover task-relevant information lost during discretization.

\begin{figure*}[!t]
    \centering
    \includegraphics[width=.85\textwidth]{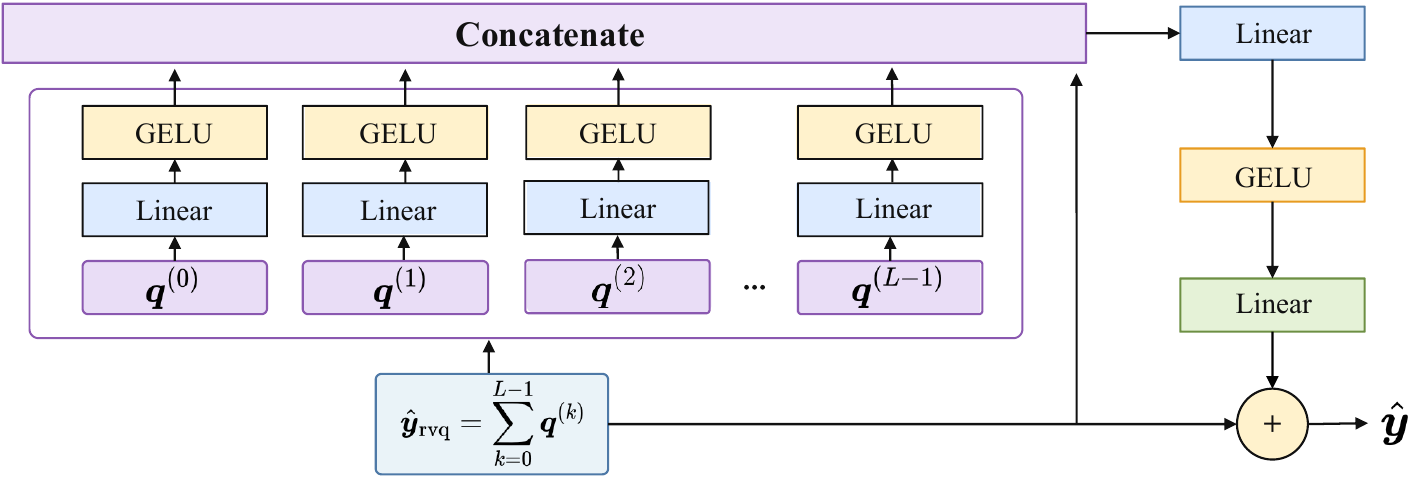}
    \caption{Architecture of the quantization error compensation adapter. The projected RVQ-layer codewords are concatenated with $\hat{\boldsymbol{y}}_{\rm rvq}$ to predict a residual correction $\boldsymbol{\Delta}$, which is added to $\hat{\boldsymbol{y}}_{\rm rvq}$ to obtain the refined feature.}
    \label{fig:qeca}
\end{figure*}

As illustrated in Fig.~\ref{fig:qeca}, QECA consists of two components:
\begin{enumerate}
    \item \emph{Per-layer projections:} each codeword $\boldsymbol{q}^{(k)}$ is independently projected through a lightweight network $g_k(\cdot) = \operatorname{GELU}(\boldsymbol{W}_k \boldsymbol{q}^{(k)})$ that maps the $d_{\rm v}$-dimensional codeword to a hidden representation of dimension $d_{\rm h}$.
    \item \emph{Shared compensation MLP:} the projected representations from all layers are concatenated with the STE-quantized vector $\hat{\boldsymbol{y}}_{\rm rvq}$ and fed into a two-layer MLP:
    \begin{equation}
        \boldsymbol{\Delta} = \operatorname{MLP}\!\Bigl(\bigl[\,g_0(\boldsymbol{q}^{(0)});\;\cdots;\;g_{L-1}(\boldsymbol{q}^{(L-1)});\;\hat{\boldsymbol{y}}_{\rm rvq}\,\bigr]\Bigr),
        \label{eq:qeca}
    \end{equation}
    where $[\,\cdot\,;\,\cdot\,]$ denotes concatenation along the feature dimension, and the MLP consists of a linear layer $\to$ GELU $\to$ linear layer.
\end{enumerate}
The final reconstructed feature is
\begin{equation}
    \hat{\boldsymbol{y}} = \hat{\boldsymbol{y}}_{\rm rvq} + \boldsymbol{\Delta}.
    \label{eq:recon}
\end{equation}

The last linear layer of the compensation MLP is initialized with zero weights and biases, ensuring that QECA initially produces no residual correction (i.e., $\boldsymbol{\Delta} = \boldsymbol{0}$).
This initialization prevents the adapter from perturbing the initial quantized features and improves stability during early training.

QECA is further optimized through the next-token prediction loss together with the trainable LoRA adapters in both the vision encoder and the LLM (see Section~\ref{sec:training}). Therefore, its predicted residual is guided by downstream answer generation rather than by feature-level reconstruction objectives alone. This task-driven training enables QECA to compensate for RVQ-induced distortion in a manner aligned with the downstream inference objective.

\subsection{Training Strategy}
\label{sec:training}

The overall training proceeds in two stages.

\emph{Stage I --- RVQ and QECA pretraining.}
The RVQ codebooks and QECA are first pretrained on visual features extracted from the training images, without involving the LLM or textual instructions.
The training objective combines cosine similarity loss and L2 normalization loss between the reconstructed and original features, along with a commitment loss that encourages the input vectors to stay close to their assigned codewords:
\begin{equation}
    \mathcal{L}_{\rm vq} = \lambda_1 \bigl(1 - \operatorname{cos}(\hat{\boldsymbol{y}},\, \boldsymbol{y})\bigr) + \lambda_2 \|\hat{\boldsymbol{y}}_{\rm norm} - \boldsymbol{y}_{\rm norm}\|^2 + \lambda_3 \mathcal{L}_{\rm commit},
    \label{eq:vq_loss}
\end{equation}
where $\mathcal{L}_{\rm commit} = \frac{1}{L}\sum_{k=0}^{L-1} \|\boldsymbol{r}^{(k)} - \operatorname{sg}[\boldsymbol{q}^{(k)}]\|^2$ is the commitment loss, $\operatorname{sg}[\cdot]$ denotes the stop-gradient operator, and $\lambda_1, \lambda_2, \lambda_3$ are weighting coefficients.

\emph{Stage II --- End-to-end task fine-tuning.}
After feature-level pretraining, we integrate the RVQ codec, multimodal projector, query-aware scoring module, and QECA with the pretrained vision encoder and LLM. Low-rank adaptation (LoRA) adapters are attached to both the vision encoder and the LLM \cite{lora}.
The visual-side adapters allow the feature distribution to adapt to the discrete RVQ bottleneck under the joint task and VQ objectives, while retaining the pretrained vision-encoder weights.
The trainable components are jointly optimized on an instruction-tuning dataset using the standard autoregressive language modeling loss:
\begin{equation}
    \mathcal{L}_{\rm lm} = -\sum_{t=1}^{T} \log p_\theta(a_t \mid \boldsymbol{x}_{\rm vis}, \boldsymbol{x}_{\rm txt}, a_{<t}),
    \label{eq:lm_loss}
\end{equation}
where $a_t$ is the $t$-th token of the target answer, $\boldsymbol{x}_{\rm vis}$ and $\boldsymbol{x}_{\rm txt}$ denote the visual and textual token sequences, respectively, and $\theta$ comprises the multimodal projector, query-aware scoring module, QECA, and the LoRA parameters in both the vision encoder and the LLM. The pretrained base weights of both the vision encoder and the LLM remain frozen, while their attached LoRA parameters are updated. The RVQ codebooks continue to be updated by EMA during this stage.

The VQ loss $\mathcal{L}_{\rm vq}$ from \eqref{eq:vq_loss} is added as a regularization term to prevent the reconstructed features from drifting too far from the originals:
\begin{equation}
    \mathcal{L} = \mathcal{L}_{\rm lm} + \beta \cdot \mathcal{L}_{\rm vq}.
    \label{eq:total_loss}
\end{equation}

\subsection{Complexity Analysis}
\label{sec:complexity}

On the device, the additional operations introduced by Q-TOFC are query-aware scoring, DPC-KNN feature merging, and RVQ index generation.
Computing the visual--text similarity matrix scales as
\begin{equation}
    \mathcal{O}(n_{\rm p} n_{\rm v} N_{\rm txt} d_{\rm s}).
    \label{eq:scoring_complexity}
\end{equation}
where $d_{\rm s}$ is the dimension of the shared relevance space.
DPC-KNN computes pairwise feature distances within each patch, giving $\mathcal{O}(n_{\rm p}n_{\rm v}^2d_{\rm v})$, while RVQ nearest-codeword search scales as
\begin{equation}
    \mathcal{O}\!\left(n_{\rm p} n_{\rm c} d_{\rm v}\sum_{k=0}^{L-1} M_k\right),
    \label{eq:rvq_complexity}
\end{equation}
where $M_k$ is the size of the $k$-th codebook. On the server, codebook lookup and summation cost $\mathcal{O}(n_{\rm p}n_{\rm c}Ld_{\rm v})$, followed by a QECA operation whose cost is linear in the number of merged features. The measured device-side encoding and end-to-end costs of these operations are reported in Section~\ref{sec:experiment}.

\section{Experimental Results}
\label{sec:experiment}

\subsection{Experimental Setup}
\label{sec:exp_setup}

\textbf{Backbone model.}
We use LLaVA-OneVision-7B \cite{llava-ov} as the backbone LMM and SigLIP-SO400M \cite{SigLIP} as its vision encoder.
Each $384 \times 384$ image patch yields $n_{\rm v} = 729$ visual features of dimension $d_{\rm v} = 1152$.
When represented in FP16, i.e., $b_{\rm f}=16$, the uncompressed features of one patch require approximately 1.60\,MiB.

\textbf{Q-TOFC configuration.}
Unless otherwise stated, we set the number of RVQ layers to $L = 8$, the first-stage codebook size to $M_0 = 16{,}384$, and $M_k = 4{,}096$ for the subsequent layers.
The RVQ indices therefore require $B_{\rm tok} = 14 + 7 \times 12 = 98$ bits per merged feature.
We vary $n_{\rm c}$ to obtain the rate--performance curves. The latency comparison and RVQ-depth ablation use $n_{\rm c}=64$.
Because LLaVA-OneVision uses a variable number of image patches, we report average per-request communication cost in KiB for each benchmark.
The QECA hidden dimension is $d_{\rm h} = 256$.
The text guidance strength is $\alpha = 0.5$ by default, and the effect of disabling query-aware scoring is studied in the ablation experiments.

\textbf{Training.}
Both stages use the LLaVA-1.5 second-stage instruction-tuning dataset \cite{llava1.5}. In Stage~I, only the images from this dataset are used: visual features are extracted to pretrain the RVQ codebooks and QECA without involving the LMM or textual instructions. In Stage~II, all 665K instruction-tuning samples are used for task-driven fine-tuning. The RVQ codebooks are updated by EMA, while the multimodal projector, query-aware scoring module, QECA, and LoRA adapters in both the SigLIP vision encoder and the LLM are optimized by backpropagation. The LoRA rank is set to 64 for both the SigLIP vision encoder and the LLM, and Stage~II training is performed for one epoch. We set
$(\lambda_1,\lambda_2,\lambda_3)=(1.5,0.5,0.05)$ in \eqref{eq:vq_loss} and $\beta=0.15$ in \eqref{eq:total_loss}.

\textbf{Evaluation benchmarks.}
Following the benchmark suite in \cite{TOFC}, we evaluate on seven multimodal benchmarks: RealWorldQA, MME \cite{MME}, AI2D \cite{AI2D}, MMBench \cite{MMB}, MMStar \cite{MMStar}, ScienceQA \cite{ScienceQA}, and MMMU \cite{MMMU}.
All evaluations are conducted using the VLMEvalKit framework \cite{vlmevalkit} to ensure a fair comparison.
These benchmarks cover complementary forms of visual understanding rather than a single narrow task.
RealWorldQA evaluates question answering grounded in real-world images, while MME covers a broad range of multimodal perception abilities.
MMBench uses multiple-choice questions to assess diverse multimodal capabilities, whereas MMStar emphasizes samples that require visual information across six core capabilities.
AI2D evaluates question answering over scientific diagrams, ScienceQA contains multimodal multiple-choice science questions, and MMMU requires college-level reasoning across multiple disciplines and heterogeneous visual inputs.
The seven-benchmark average therefore aggregates performance across real-world scenes, diagrams, scientific questions, and knowledge-intensive reasoning tasks.

\textbf{Baselines.}
We compare Q-TOFC against the following compression methods:
\begin{itemize}
    \item \textbf{TOFC} \cite{TOFC}: Clustering-based feature merging with learnable entropy coding and the closest feature-level baseline.
    \item \textbf{JPEG} \cite{wallace1991jpeg}: Standard image compression at varying quality levels.
    \item \textbf{ELIC} \cite{ELIC}: Neural image compression with autoregressive context modeling.
\end{itemize}
For image-level baselines, the compressed image is decoded and re-encoded by the vision encoder at the edge server. The resulting visual features are then merged using DPC-KNN to produce 128 merged features per patch, thereby controlling the visual-token workload of subsequent LLM inference. All methods use the same LMM backbone and evaluation framework, and the runtime of each operation is included on the side where it is executed.

\textbf{Score aggregation and communication accounting.}
For each benchmark, we first compute the score of a compressed method using the benchmark-specific evaluation metric implemented in VLMEvalKit.
The score is then normalized by the score of the uncompressed LLaVA-OneVision-7B backbone on the same benchmark.
The average score reported in Fig.~\ref{fig:avg_rate} is the arithmetic mean of these normalized scores across the seven benchmarks.
This normalization prevents a benchmark with a numerically larger metric range from dominating the average.

Communication cost is reported as the visual payload transmitted per request in KiB. The textual query is excluded because it is identical across all compared methods. To obtain the rate--performance curves, we vary the number of merged features $n_{\rm c}$ for Q-TOFC while keeping its RVQ configuration fixed. For TOFC, we train separate models with different rate--distortion loss weights. The ELIC operating points use the official pretrained checkpoints corresponding to different rate--distortion settings, while the JPEG operating points are generated by varying the quality factor. All resulting outputs are evaluated using the same benchmark scripts.

\subsection{Rate--Performance Trade-off}

\begin{figure}[!t]
    \centering
    \includegraphics[width=1\linewidth]{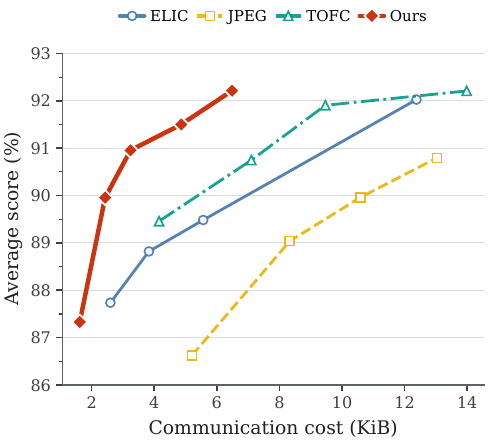}
    \caption{Communication--performance trade-off averaged over seven benchmarks with LLaVA-OneVision-7B. Each benchmark score is normalized by the corresponding uncompressed LLaVA-OneVision-7B score before averaging.}
    \label{fig:avg_rate}
\end{figure}

Fig.~\ref{fig:avg_rate} shows the communication--performance trade-off obtained by sweeping the compression settings of each method.
Q-TOFC occupies the low-communication region of the curve: even at 3.24\,KiB it preserves 90.95\% of the backbone's average normalized score, and at 6.49\,KiB it reaches 92.21\%.
The latter operating point essentially matches the 92.20\% score of TOFC at 13.98\,KiB while requiring only 46.4\% of its payload.
Across the plotted operating points, Q-TOFC also lies above the image-level JPEG and ELIC baselines in the low-communication region.
Among the image-level baselines, ELIC consistently outperforms JPEG, demonstrating the benefit of learned neural image compression over conventional image coding under the evaluated settings.
However, ELIC remains below the feature-level methods at comparable communication costs, and JPEG does not reach their higher-performance range within the displayed payloads.
This comparison indicates that improving the image codec narrows the performance gap, but directly transmitting task-oriented visual features remains more effective for downstream LMM inference under limited communication budgets.
Increasing the number of merged features steadily improves Q-TOFC performance, confirming that retaining more visual evidence benefits multimodal reasoning.
The curve becomes flatter at the higher-rate operating points, however, indicating diminishing returns from further increasing the number of merged features under the fixed 98-bit representation.
Thus, varying the number of merged features provides a direct control knob: smaller values favor communication efficiency, whereas larger values recover additional task performance.

\begin{figure}[!t]
    \centering
    \includegraphics[width=1\linewidth]{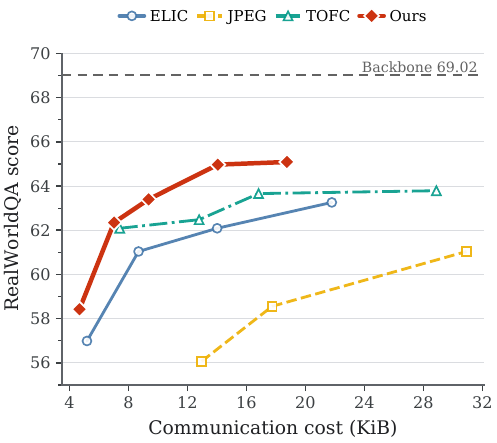}
    \caption{Communication--performance trade-off on RealWorldQA with LLaVA-OneVision-7B. Communication cost is the average visual payload transmitted per request, and the dashed line indicates the uncompressed-backbone score.}
    \label{fig:rwqa_rate}
\end{figure}

Fig.~\ref{fig:rwqa_rate} presents the communication--performance trade-off on RealWorldQA.
At a comparable accuracy level, Q-TOFC reduces the visual payload by 23.7\% relative to TOFC.
This result is consistent with the design of preserving a larger set of compact, query-relevant merged features instead of aggressively reducing the token count.
The larger separation from JPEG is consistent with the sensitivity of fine edges and text strokes to pixel-domain compression.
ELIC narrows this gap through learned image compression, but both image-level baselines remain below the feature-level methods on this benchmark, indicating the value of feature-space transmission for preserving task-relevant visual information.
The qualitative examples in Section~\ref{sec:query_case} examine this behavior for localized sign evidence at the cluster-center level.

\begin{figure}[!t]
    \centering
    \includegraphics[width=1\linewidth]{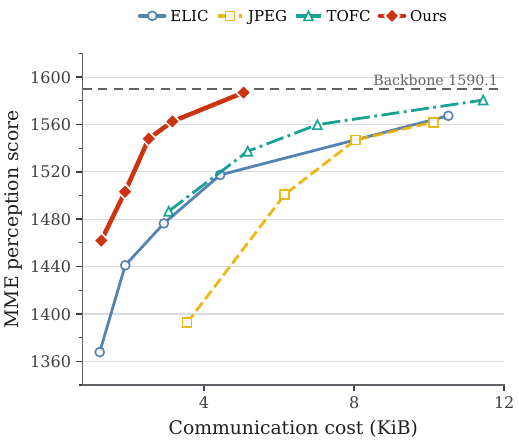}
    \caption{Communication--performance trade-off on MME with LLaVA-OneVision-7B. Communication cost is the average visual payload transmitted per request, and the dashed line indicates the uncompressed-backbone score.}
    \label{fig:mme_rate}
\end{figure}

Fig.~\ref{fig:mme_rate} shows the communication--performance results on MME.
Q-TOFC rises rapidly in the low-payload region and approaches the uncompressed backbone performance with only a few KiB per request.
At its 5.05\,KiB operating point, Q-TOFC surpasses the highest plotted scores of TOFC, ELIC, and JPEG while reducing the communication payload by 55.8\%, 51.9\%, and 66.8\%, respectively.
Among the baselines, TOFC achieves a better communication--performance trade-off than ELIC and JPEG, demonstrating the advantage of directly coding task-oriented visual features over transmitting compressed images.
Together with the RealWorldQA results, the MME curve shows that Q-TOFC maintains its communication--performance advantage across benchmarks with different visual information requirements.
\subsection{Latency Analysis}

\begin{figure*}[!t]
    \centering
    \includegraphics[width=.92\textwidth]{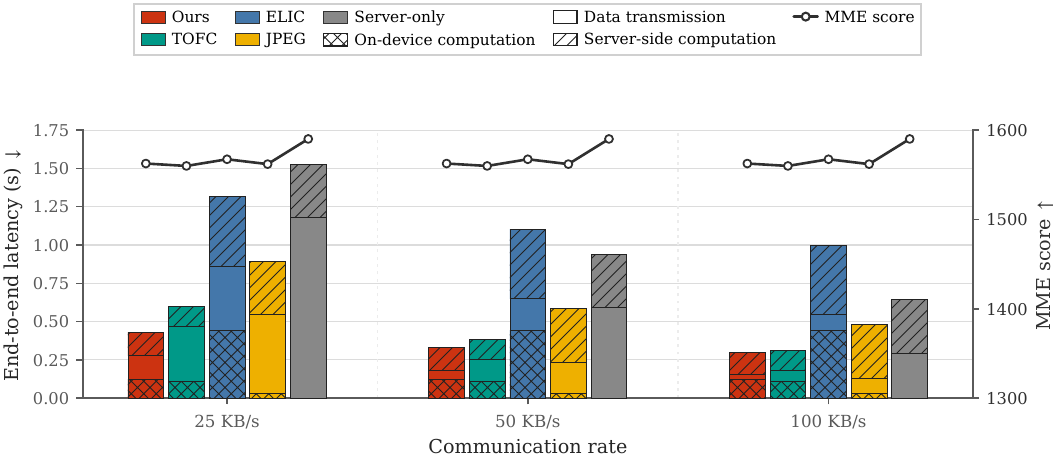}
    \caption{End-to-end latency breakdown and MME scores with LLaVA-OneVision-7B at uplink rates of 25, 50, and 100\,KB/s. Stacked bars show on-device computation, data transmission, and server-side computation, while markers indicate the corresponding MME scores. Q-TOFC and TOFC use $n_{\rm c}=64$ and $n_{\rm c}=16$, respectively; JPEG uses $q=20$, and ELIC uses $\lambda=0.15$.}
    \label{fig:latency}
\end{figure*}

\begin{figure}[!t]
    \centering
    \includegraphics[width=1\linewidth]{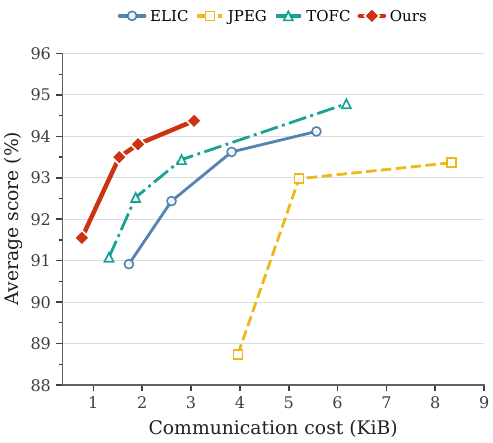}
    \caption{Communication--performance trade-off averaged over seven benchmarks with LLaVA-1.5-7B. Each benchmark score is normalized by the corresponding uncompressed LLaVA-1.5-7B score before averaging.}
    \label{fig:avg_rate_llava15}
\end{figure}

Fig.~\ref{fig:latency} breaks down the end-to-end latency under uplink rates of 25, 50, and 100\,KB/s.
For the latency comparison, we select operating points that yield similar MME scores across the compressed methods.
The transmission latency is computed from the visual payload and the assumed uplink rate, while the device-side and server-side components are measured computation times.
The device-side measurements use an NVIDIA Jetson AGX Orin, and the edge-side measurements use NVIDIA RTX 4090 hardware.
All computation and end-to-end latency results are averaged over the evaluation requests.

Q-TOFC and TOFC have comparable device-side processing times, while Q-TOFC transmits a smaller payload and therefore requires substantially less transmission time.
At 25\,KB/s on MME, Q-TOFC lowers the end-to-end latency by 29.0\%, 67.6\%, and 54.2\% relative to TOFC, ELIC, and JPEG, respectively.
The comparison includes all method-specific processing operations: JPEG and ELIC perform image decoding and visual feature extraction at the edge server, whereas TOFC and Q-TOFC compress intermediate visual features on the device and reconstruct them at the server.
All these operations are included in the reported end-to-end latency.
For a fixed Q-TOFC configuration, each image patch produces the same number of RVQ index bits.
Once the patch count is known, the request payload can therefore be determined before transmission, which may simplify uplink resource allocation and latency-aware scheduling in practical deployments.
As the uplink rate increases, transmission accounts for a smaller fraction of the total latency, and the latency gap between Q-TOFC and TOFC consequently narrows.
Nevertheless, Q-TOFC maintains the lowest end-to-end latency among the compressed methods across all evaluated uplink rates.
The score markers in Fig.~\ref{fig:latency} confirm that this latency advantage is achieved while maintaining similar MME performance.

\subsection{Extension to LLaVA-1.5}

\begin{figure*}[!t]
    \centering
    \includegraphics[width=.92\textwidth]{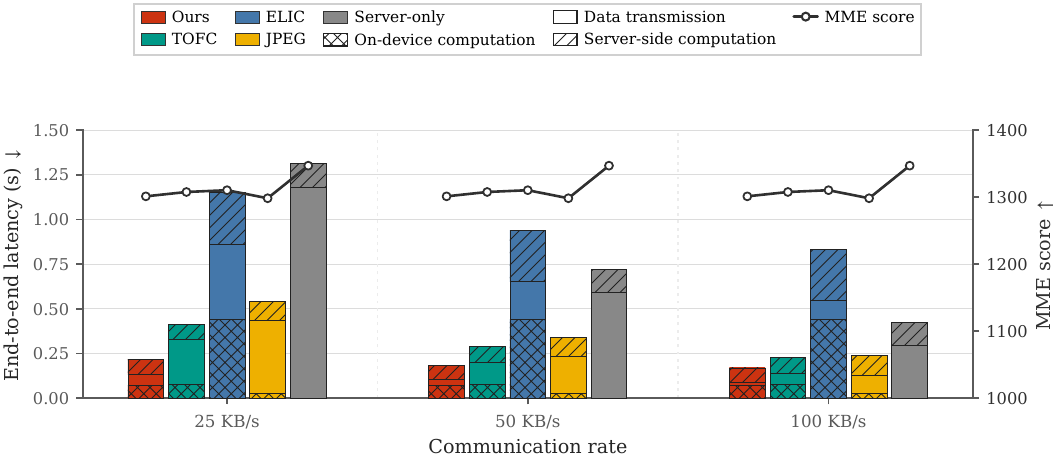}
    \caption{End-to-end latency breakdown and MME scores with LLaVA-1.5-7B at uplink rates of 25, 50, and 100\,KB/s. Stacked bars show on-device computation, data transmission, and server-side computation, while markers indicate the corresponding MME scores. Q-TOFC and TOFC use $n_{\rm c}=64$ and $n_{\rm c}=32$, respectively; JPEG uses $q=20$, and ELIC uses $\lambda=0.32$.}
    \label{fig:latency_llava15}
\end{figure*}

To examine whether Q-TOFC generalizes beyond the primary LLaVA-OneVision backbone, we further evaluate it with LLaVA-1.5-7B \cite{llava1.5}, which uses a CLIP ViT-L/14-336 vision encoder \cite{CLIP}.
Each input image is resized and padded to $336 \times 336$ pixels and processed as a single image patch, yielding $n_{\rm v}=576$ visual features of dimension $d_{\rm v}=1024$.
We vary the number of merged features from $n_{\rm c}=64$ to 256 and train a separate Q-TOFC instance using the same two-stage procedure described in Section~\ref{sec:exp_setup}. The remaining compression and evaluation settings are unchanged. For the average score, each benchmark is normalized by the corresponding uncompressed LLaVA-1.5 result.

Fig.~\ref{fig:avg_rate_llava15} shows that the communication advantage of Q-TOFC persists with the CLIP-based backbone.
At comparable average normalized performance, Q-TOFC reduces the payload by 45.4\%, 60.0\%, and 81.6\% relative to the closest-performance operating points of TOFC, ELIC, and JPEG, respectively.
At the higher-performance operating points, Q-TOFC requires approximately half the payload of TOFC while maintaining similar performance.
The absolute payloads are lower than those with LLaVA-OneVision because LLaVA-1.5 processes a single fixed-resolution image patch rather than a variable number of patches.

Fig.~\ref{fig:latency_llava15} reports the MME latency results with LLaVA-1.5-7B.
Q-TOFC achieves the lowest end-to-end latency among the compressed methods at all three uplink rates while maintaining comparable MME performance.
At 25\,KB/s, Q-TOFC reduces the end-to-end latency by 48.2\%, 81.3\%, and 60.3\% relative to TOFC, ELIC, and JPEG, respectively.
Q-TOFC also maintains the lowest latency at the other evaluated uplink rates, while the score markers show that the compared methods achieve similar MME performance.
Together with Fig.~\ref{fig:avg_rate_llava15}, these results show that Q-TOFC retains its communication and latency advantages when the vision encoder changes from SigLIP to CLIP and the image representation changes from variable-patch to fixed-resolution input.

\subsection{Ablation Studies}

\begin{figure}[!t]
    \centering
    \includegraphics[width=1\linewidth]{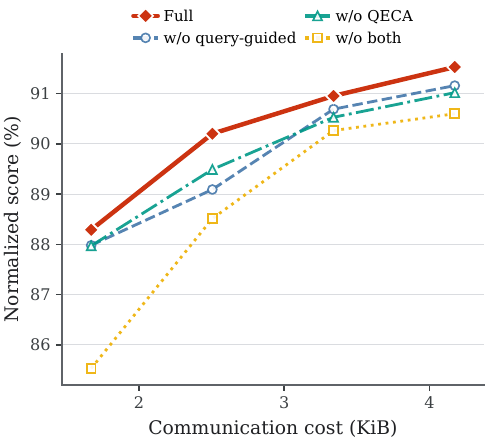}
    \caption{Ablation of query-aware token scoring and QECA with LLaVA-OneVision. Scores are averaged over seven benchmarks after normalization by the corresponding uncompressed-backbone scores. At each operating point, all variants use the same RVQ configuration and merged-feature count.}
    \label{fig:ablation}
\end{figure}

\begin{figure*}[!t]
    \centering
    \includegraphics[width=.495\linewidth]{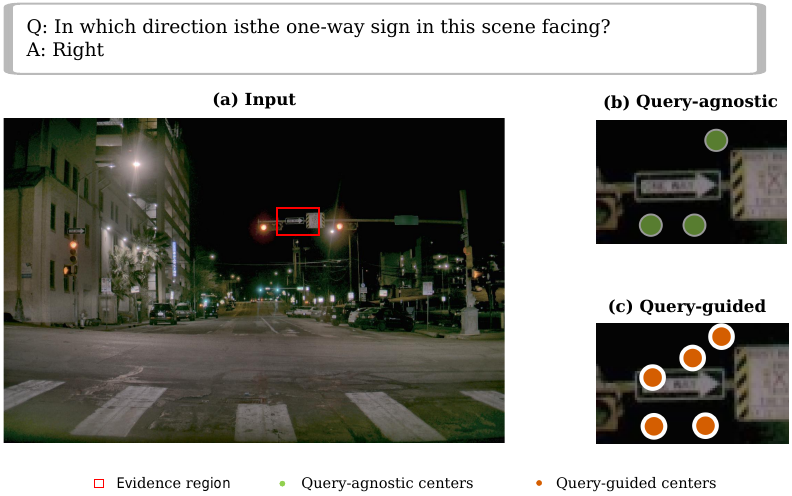}\hfill
    \includegraphics[width=.495\linewidth]{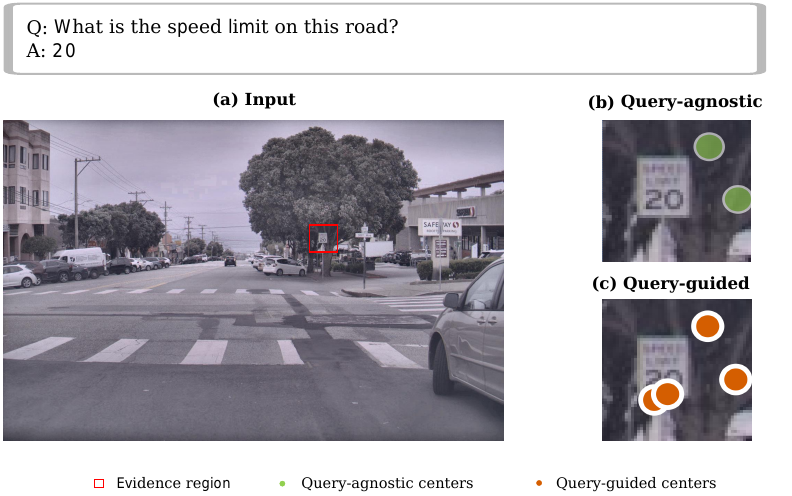}
    \caption{Qualitative examples of query-guided cluster-center selection on RealWorldQA. With identical visual features and the same total number of selected centers, query guidance increases the number of centers in the displayed evidence region from three to five (left) and from two to four (right). The markers denote visual-token locations projected onto the image grid rather than pixel-level attention.}
    \label{fig:query_case}
\end{figure*}

\begin{table*}[!t]
\centering
\caption{Effect of RVQ depth under the 64-feature setting.}
\label{tab:rvq_depth}
\footnotesize
\setlength{\tabcolsep}{3.0pt}
\renewcommand{\arraystretch}{1.08}
\resizebox{.90\textwidth}{!}{%
\begin{tabular}{|l|c|c|c|c|c|c|c|c|c|c|c|}
\hline
\textbf{Model} & \textbf{$L$} & \textbf{$B_{\rm tok}$} & \textbf{CB storage (MB)} & \textbf{MME} & \textbf{MMB} & \textbf{MMStar} & \textbf{AI2D} & \textbf{RWQA} & \textbf{SQA} & \textbf{MMMU} & \textbf{Avg. norm} \\
\hline
Backbone & -- & -- & -- & 1590.10 & 82.13 & 61.87 & 82.70 & 69.02 & 95.34 & 48.00 & 100.00\% \\
\hline
Q-TOFC & 2 & 26 & 47.19 & 1498.72 & 73.53 & 46.07 & 72.18 & 61.57 & 80.54 & 44.00 & 87.27\% \\
\hline
Q-TOFC & 4 & 50 & 66.06 & 1509.10 & 76.54 & 47.93 & 72.11 & 63.27 & 82.50 & 44.67 & 89.15\% \\
\hline
Q-TOFC & 6 & 74 & 84.93 & 1527.44 & 76.97 & 48.80 & 73.21 & 63.79 & 83.40 & 45.22 & 90.18\% \\
\hline
Q-TOFC & 8 & 98 & 103.81 & 1557.90 & 76.89 & 49.13 & 73.48 & 63.40 & 86.17 & 45.40 & 90.95\% \\
\hline
\end{tabular}
}
\end{table*}

Fig.~\ref{fig:ablation} evaluates the contributions of query-aware scoring and QECA under different merged-feature counts.
The full configuration consistently achieves the highest normalized score across the entire communication--performance curve, indicating that the gains are not limited to a single operating point.
Removing either query-aware scoring or QECA shifts the curve downward, while disabling both modules produces the largest performance degradation.
The separation is more visible in the low-payload region and becomes smaller as more merged features are retained.
This trend suggests that both modules are particularly useful when the compressed representation has limited capacity.

The two modules improve different stages of the compression pipeline.
Query-aware scoring affects feature selection before quantization by increasing the likelihood that query-relevant visual features are retained during merging.
QECA operates after RVQ reconstruction and refines the quantized representations before they are processed by the LMM.
Removing both modules therefore affects both feature selection and feature reconstruction.
The consistently stronger performance of the full configuration supports using query-aware scoring and QECA together in Q-TOFC.

Table~\ref{tab:rvq_depth} further studies the effect of RVQ depth while fixing the number of merged features per patch to 64 and keeping both query-guided clustering and QECA enabled.
For an $L$-layer RVQ, each merged feature requires $B_{\rm tok}=14+(L-1)\times 12$ index bits under our codebook configuration.
Increasing RVQ depth therefore raises both the number of index bits per merged feature and the device-side codebook storage, which grows from 47.19\,MB at $L=2$ to 103.81\,MB at $L=8$ when the codebooks are stored in FP16.
Increasing the number of RVQ codebooks improves the average normalized score from 87.27\% at $L=2$ to 90.95\% at $L=8$, showing that residual refinement is important for preserving task-relevant visual features.
Most of this improvement is obtained by the first six layers: increasing the depth from $L=6$ to $L=8$ adds 24 bits per merged feature and 18.88\,MB of codebook storage, while improving the average normalized score by 0.77 percentage points.
This diminishing return reveals a three-way trade-off among communication cost, codebook memory, and task performance.
We use $L=8$ as the default high-accuracy configuration, while the $L=6$ result provides a lower-rate, lower-memory alternative.

\subsection{Qualitative Analysis of Query-Guided Center Selection}
\label{sec:query_case}

To complement the aggregate ablation results, Fig.~\ref{fig:query_case} visualizes how query guidance changes cluster-center selection in two RealWorldQA examples. For each example, the query-agnostic and query-guided variants use the same visual features and select the same total number of cluster centers. The only difference is whether the normalized text--visual relevance score is applied to the DPC-KNN center-ranking criterion. The selected visual-token locations are projected onto the input-image grid. For readability, the figure enlarges the evidence region required by the question and displays only the centers within the same crop for both variants. These markers indicate projected visual-token locations rather than pixel-level attention.

In the first example, the question asks, ``In which direction is the one-way sign in this scene facing?'' Query guidance increases the number of selected centers in the displayed sign region from three to five, thereby allocating more representation capacity to evidence relevant to recognizing its direction. Without query guidance, the model answers ``left.'' With query-guided center selection, the answer changes to ``right,'' which matches the ground truth. In the second example, the question asks, ``What is the speed limit on this road?'' The number of centers in the displayed speed-limit-sign region increases from two to four. Correspondingly, the query-agnostic variant answers ``25,'' whereas the query-guided variant answers ``20,'' which is correct. Since the total number of selected centers is unchanged in both comparisons, these changes represent a reallocation toward the localized evidence identified by the query rather than an increase in the total number of centers. Although qualitative, the two cases connect this reallocation with corrected downstream predictions and illustrate the mechanism underlying the quantitative gains observed in Fig.~\ref{fig:ablation}.

\FloatBarrier

\section{Conclusion}
\label{sec:conclusion}

This paper presented Q-TOFC, a task-oriented visual feature compression framework for device--edge collaborative multimodal inference.
Q-TOFC uses query-aware feature scoring to guide cluster-center selection toward visual evidence relevant to the user's textual query.
It employs RVQ to represent each merged visual feature as a compact sequence of codebook indices, allowing more merged features to be retained at low communication cost.
To mitigate the distortion introduced by multi-layer quantization, QECA refines the reconstructed features under the joint supervision of the task objective and feature-level VQ regularization.
Experiments on seven multimodal benchmarks with LLaVA-OneVision showed that Q-TOFC reduces the visual payload by 53.6\% relative to TOFC while maintaining comparable average normalized task performance.
The lower payload further translates into reduced end-to-end latency under bandwidth-constrained uplinks.
Additional experiments with LLaVA-1.5 demonstrate that these communication and latency advantages persist with a different vision encoder and image-patching scheme.
Ablation studies confirm that query-aware feature scoring and QECA address complementary sources of information loss, while the RVQ-depth study characterizes the trade-off among communication cost, codebook storage, and task performance.
Future work will investigate dynamic selection of both RVQ depth and the number of merged features according to channel conditions.

\FloatBarrier

\bibliographystyle{IEEEtran}
\bibliography{refs}

\end{document}